\documentclass[conference,a4paper]{IEEEtran}
\usepackage{stmaryrd}
\usepackage{amsfonts}

\usepackage{fancyhdr}
\usepackage{graphicx,times,amsmath} 

\IEEEoverridecommandlockouts 

\usepackage{bm}
\usepackage{subfigure}
\usepackage{algorithm}
\usepackage{algorithmic}
\usepackage{multirow}
\usepackage{makecell}

\graphicspath{{figures/}}

\begin{document}

\title{\LARGE\bf A New Manifold Distance Measure for Visual Object Categorization
\thanks{F. Li, X. Huang and B. Zhang are with Institute of Applied Mathematics,
AMSS, Chinese Academy of Sciences, Beijing 100190, China
(email: \{lifengfu12,huangxiayuan11\}@mails.ucas.ac.cn, b.zhang@amt.ac.cn).}
\thanks{H. Qiao is with the State Key Lab of Management and Control for Complex Systems,
Institute of Automation, Chinese Academy of Sciences, Beijing 100190, China
and CAS Center for Excellence in Brain Science and Intelligence Technology (CEBSIT),
Shanghai 200031, China (email: hong.qiao@ia.ac.cn)}
\thanks{This work was partly supported by NSFC under Grants No. 61210009, 61503383, 61379093 and 11131006
and the Strategic Priority Research Program of CAS (Grant No. XDB02080003),
and Beijing Natural Science Foundation under Grant No. 2141100002014002.}
}
\author{Fengfu Li, Xiayuan Huang, Hong Qiao and Bo Zhang}

\maketitle

\begin{abstract}
Manifold distances are very effective tools for visual object recognition. However, most of
the traditional manifold distances between images are based on the pixel-level comparison and
thus easily affected by image rotations and translations. In this paper, we propose a new manifold
distance to model the dissimilarities between visual objects based on the Complex Wavelet Structural
Similarity (CW-SSIM) index. The proposed distance is more robust to rotations and translations of
images than the traditional manifold distance and the CW-SSIM index based distance. In addition,
the proposed distance is combined with the $k$-medoids clustering method to derive a new clustering
method for visual object categorization. Experiments on Coil-20, Coil-100 and Olivetti Face Databases
show that the proposed distance measure is better for visual object categorization than both the
traditional manifold distances and the CW-SSIM index based distances.
\end{abstract}

\section{Introduction}

\PARstart{C}{omparing} the similarity of two objects is a fundamental operation for many clustering
algorithms \cite{bib:simone1999}. Methods such as the $k$-medoids \cite{bib:hae-sang2009},
the $dip$-means \cite{bib:argyris2012} and clustering by fast search and find of density peaks \cite{bib:alex2014}
only need the similarity matrix as input. A similarity measure is a real-world function that assesses the
similarity between two objects. Although no single definition of a similarity measure exists,
similarity measures are usually in some sense the opposite of distance metrics, that is, they take on
large values for similar objects and small values for very dissimilar objects.

There have been considerable efforts in searching for the appropriate similarity measures for object categorization.
In Euclidean space, the Euclidean distance ($L_2$ norm) and the city-block distance ($L_1$ norm) are two most famous distances. For images, one of the most effective similarity measures is complex wavelet structural similarity (CW-SSIM)
index \cite{bib:sampat2009}, which is robust to small rotations and translations of images. Pearson correlation
coefficient \cite{bib:jacob2009} and joint entropy \cite{bib:chein1994} are two widely used similarity/distance
measures for probability distributions. For high-dimensional data, Radial Basis Function (RBF) kernel is a popular
choice of a distance measure. More information about similarity/distance measures can be found in \cite{bib:liu2006}.

Although the previous distance/similarity measures have achieved some success, they are not suitable to deal with
visual objects which often lie in very high-dimensional spaces and have 2D/3D spatial structures \cite{bib:pinto2008}.
On one hand, manifold learning is a powerful tool to deal with the high-dimensional issue of visual objects.
Most of the manifold learning methods use the manifold ways of perception \cite{bib:seung2000} which assumes
that the data of interest lie on an embedded low-dimensional manifold within the high-dimensional space
whose dimension equals to the size of an image. Isomap \cite{bib:tenenbaum2000}, one of the most famous manifold
learning algorithms, implements the manifold assumption by introducing the geometric distance to approximate
the manifold distance. The geometric distance itself is approximated by the shortest distance on a neighborhood
graph constructed by $\epsilon$-neighborhood or $t$-nearest-neighborhood ($t$-nn) neighborhood (here we use $t$
instead of $k$ in the $k$-nearest-neighborhood algorithm to avoid confusion with the $k$ used in $k$-means/$k$-medoids).
The advantage of the geometry distance is that it preserves the neighborhood properties of the data distribution
and keeps the intrinsic dimension unchanged. Manifold learning methods have been successfully applied into
dimensionality reduction \cite{bib:lle2000}, target tracking \cite{bib:mani-tracking2014}, face recognition \cite{bib:mani-facerecog2008} and so on.

One the other hand, the structural similarity (SSIM) index \cite{bib:ssim2004} is one of the first attempt to
deal with the 2D/3D structure of visual objects. It attempts to discount those distortions that do not affect
the structures of the image and achieves a very good performance for image quality prediction with a wide
variety of image distortions. However, it is highly sensitive to translations and rotations of images.
To address this issue, the CW-SSIM index was proposed in \cite{bib:sampat2009}, which assumed that the local
phase pattern contains more structural information than the local magnitude, and the non-structural image
distortions such as small translations lead to consistent phase shift of a group of neighboring wavelet
coefficients. The CW-SSIM index is very robust to small rotations and translations of images and can be
combined with other image clustering \cite{bib:alex2014} or classification \cite{bib:cwssim-classification2011}
methods.

Although manifold learning methods and the CW-SSIM index have achieved some success in their own fields,
no effort has been made to investigate the merit of combining the two methods. In this paper, we propose
a new manifold distance measure named the Geometric CW-SSIM (GCW-SSIM) distance measure for visual object
categorization. The distance is a combination of the CW-SSIM index for measuring the similarities between
images and the geometric distance for preserving the local properties of a cluster. In addition, we apply
the new manifold distance measure to the $k$-medoids and propose a new clustering method named the Geometric
CW-SSIM $k$-medoids (GCW-SSIM $k$-medoids). Further, experiments have also been conducted on three famous
visual object data sets, Coil-20 \cite{bib:coil20}, Coil-100 \cite{bib:coil100}, and Olivetti Face
Database \cite{bib:olivetti1994} to evaluate the performance of both the new manifold distance measure and
the GCW-SSIM $k$-medoids.

The rest of this paper is organized as follows. Section \ref{sec:gcw-ssim} introduces the geometric CW-SSIM
distance measure. The GCW-SSIM $k$-medoids clustering algorithm is described in Section \ref{sec:gcw-ssim-k-medoids}.
In Section \ref{sec:experiments}, experiments are conducted to evaluate the performance of the proposed
geometric CW-SSIM distance measure and the GCW-SSIM $k$-medoids clustering algorithm.
Conclusions are given in Section \ref{sec:conclusion}.

\section{Geometric CW-SSIM Distance} \label{sec:gcw-ssim}

\begin{figure}[!thb]
\centering
\includegraphics[width=0.8\linewidth, bb=75 250 490 660]{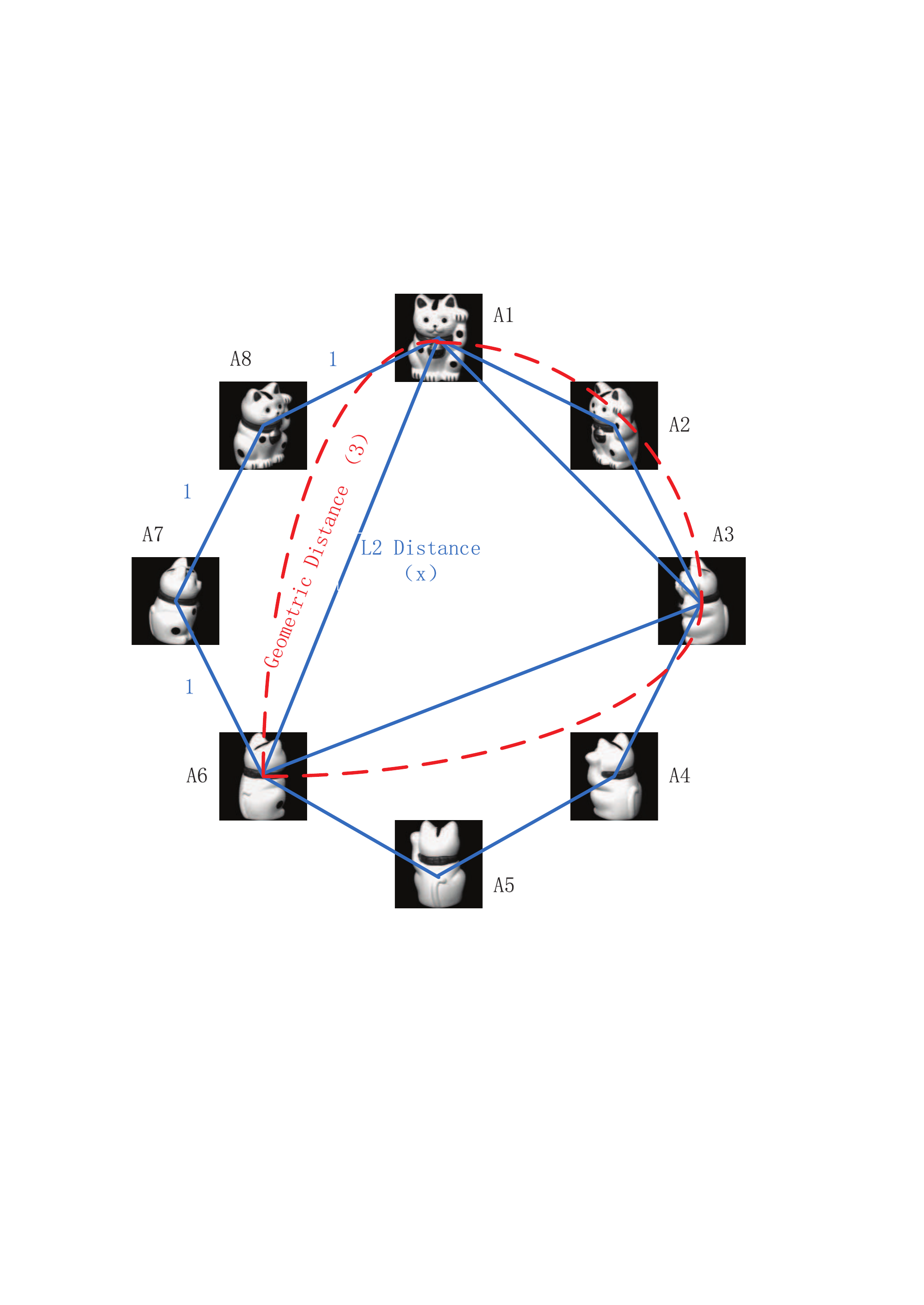}
\caption{Illustration of the difference between the Euclidean and manifold distances.
The blue lines denote the Euclidean distances, and the red dash lines denote the manifold distances.
The $L_2$ distances between two neighbors are almost the same, which we set as $1$ for simplicity.
The $L_2$ distances between other points can be any other numbers, which have nothing to do with the
distance of the neighbors. The manifold distance between A1 and A6 is approximated by their shortest
distance on the $2$-nn neighborhood graph and is about $3$ in this example.}
\label{fig:mani-demo}
\end{figure}

The idea of the geometric CW-SSIM distance first comes from the geometric distance which is a widely-used
manifold distance. Fig. \ref{fig:mani-demo} shows the difference between the traditional Euclidean distance
(or $L_2$ distance) and the manifold distance. In Fig. \ref{fig:mani-demo}, images of a toy cat taken with
different angles form a manifold space whose intrinsic dimensionality is $1.$ The reason for this is that
only the angle is a free variable. In the manifold space, it is assumed that the distance between images
is proportional to the difference between their angles (in the sense of a modulus operation of $360$ degrees).

However, the $L_2$ distance violates the manifold assumption due to the following two drawbacks:
\begin{enumerate}
\item It only considers the pixel intensities and thus neglects the geometric information between objects;
\item It neglects the $2$D structural information of images.
\end{enumerate}

One of the solutions for the first drawback is to use the geometric distance instead of the $L_2$ distance.
However, the geometry distance in the manifold is hard to estimate due to the limited and discrete samples.
An approximation for the geometric distance is the minimum distance in a neighborhood graph that is usually
constructed by $\epsilon$-neighborhood or $t$-nearest-neighborhood ($t$-nn) \cite{bib:graph-kmeans}.
The minimum distance on the neighborhood graph can be efficiently computed by the Dijkstra algorithm \cite{bib:dijkstra}.

To resolve the second drawback with the $L_2$ distance, the CW-SSIM index is used as the similarity measure
between images. The CW-SSIM index takes the $2$D structure of images into consideration and is a general index
for image similarity measurement. The key idea behind the CW-SSIM index is that small geometric image distortions
lead to consistent phase changes in the local wavelet coefficients and that a consistent phase shift of
the coefficient does not change the structural content of the image. Specifically, given two sets of coefficients
$\bm{c}_x$ and $\bm{c}_y$ extracted at the same spatial location in the same wavelet sub-bands of the two images
being compared, the local CW-SSIM index is defined as:
\begin{equation}
\tilde{S}(\bm{c}_x, \bm{c}_y)=\frac{2|\sum_{i=1}^M c_{x,i}c^*_{y,i}|+K}{\sum_{i=1}^M|c_{x,i}|^2 + \sum_{i=1}^M|c_{y,i}|^2+K}.
\end{equation}
Here, $c^*$ denotes the complex conjugate of $c$ and $K$ is a small positive stabilizing constant.
$\tilde{S}(\bm{c}_x,\bm{c}_y)$ ranges from $0$ to $1,$ where the fact that $\tilde{S}(\bm{c}_x,\bm{c}_y)$ equals $1$
implies no structural distortion. The global CW-SSIM index $\tilde{S}(\bm{x},\bm{y})$ between two images $\bm{x}$ and $\bm{y}$ is calculated as the average of $\tilde{S}(\bm{c}_x,\bm{c}_y)$, which is first computed with a sliding window
running across the whole wavelet sub-bands and then averaged. The advantage of the CW-SSIM index includes: 1) It does
not require explicit correspondences between pixels being compared; 2) It is insensitive to small geometric distortions
(rotations and translations); and 3) It compares the textural and structural properties of the localized regions of
the image pairs.

A new manifold distance called the GCW-SSIM distance is obtained by combining the CW-SSIM index and
the geometric distance. $t$-nn is used to construct the neighborhood graph. Algorithm \ref{alg:gcw-ssim}
shows the procedure for computing the GCW-SSIM distance. In Algorithm \ref{alg:gcw-ssim},
$N_t(\bm{x}^{(i)})$ is the $t$-neighborhood of $\bm{x}^{(i)}$ and Dijkstra($\bm{x^{(i)}}$, $\bm{x^{(j)}}$)
is the shortest distance between $\bm{x^{(i)}}$ and $\bm{x^{(j)}}$ on the neighborhood graph.

\begin{algorithm}[!thb]
\caption{The Geometric CW-SSIM Distance}
\label{alg:gcw-ssim}
\begin{algorithmic}[1]
\REQUIRE $\bm{X}$ and $t$ ($t$-nn parameter)
\ENSURE $\bm{G}$ (GCW-SSIM distance)
\STATE $s_{ij} \leftarrow \tilde{S}(\bm{x^{(i)}}$, $\bm{x^{(j)}}$)
\STATE $d_{ij} \leftarrow 1 - s_{ij}$
\STATE Construct the neighborhood graph using the CW-SSIM distance and $t$-nn as follows:
\begin{eqnarray}
d_{ij} \leftarrow
\begin{cases}
d_{ij}, & \text{if~} \bm{x}^{(j)} \in N_t(\bm{x}^{(i)}); \\
\infty, & \text{otherwise}.
\end{cases}
\end{eqnarray}
\STATE $g_{ij} \leftarrow $ Dijkstra($\bm{x^{(i)}}$, $\bm{x^{(j)}}$)
\RETURN $\bm{G}$
\end{algorithmic}
\end{algorithm}

Compared with the geometric distance, the GCW-SSIM distance improves the accuracy of computing
the manifold distance since the CW-SSIM index is more robust to rotations than the $L_2$ distance.
In addition, the GCW-SSIM distance is more robust to rotations and translations of images compared
with the CW-SSIM index since the manifold distance helps to preserve the local information of similar objects.

\section{The GCW-SSIM $k$-medoids} \label{sec:gcw-ssim-k-medoids}

\begin{figure*}[!thb]
\centering
\subfigure[$L_2$ distance]{\includegraphics[width=0.2\linewidth, bb=0 0 290 290]{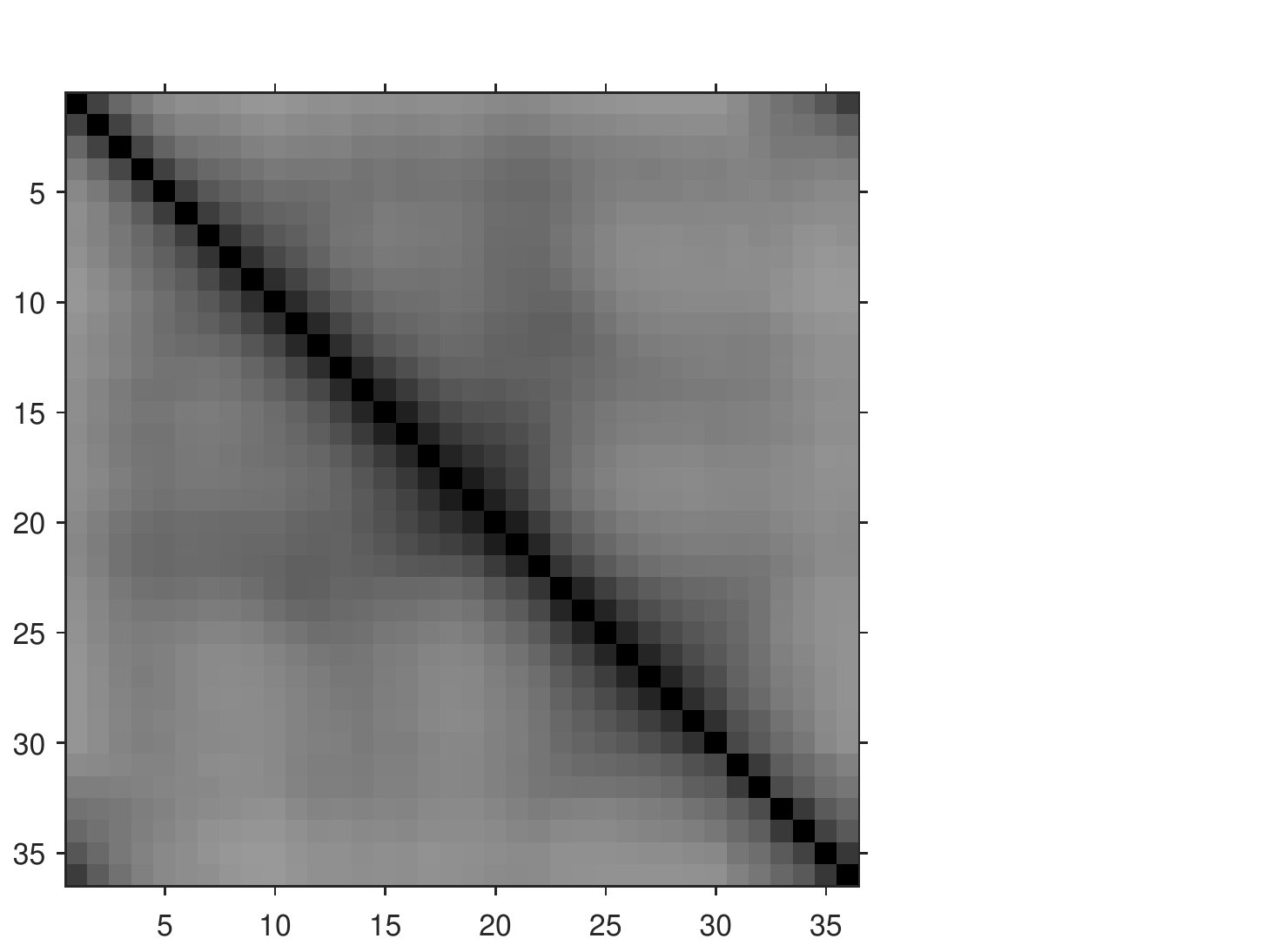}}
\subfigure[CW-SSIM distance]{\includegraphics[width=0.2\linewidth, bb=0 0 290 290]{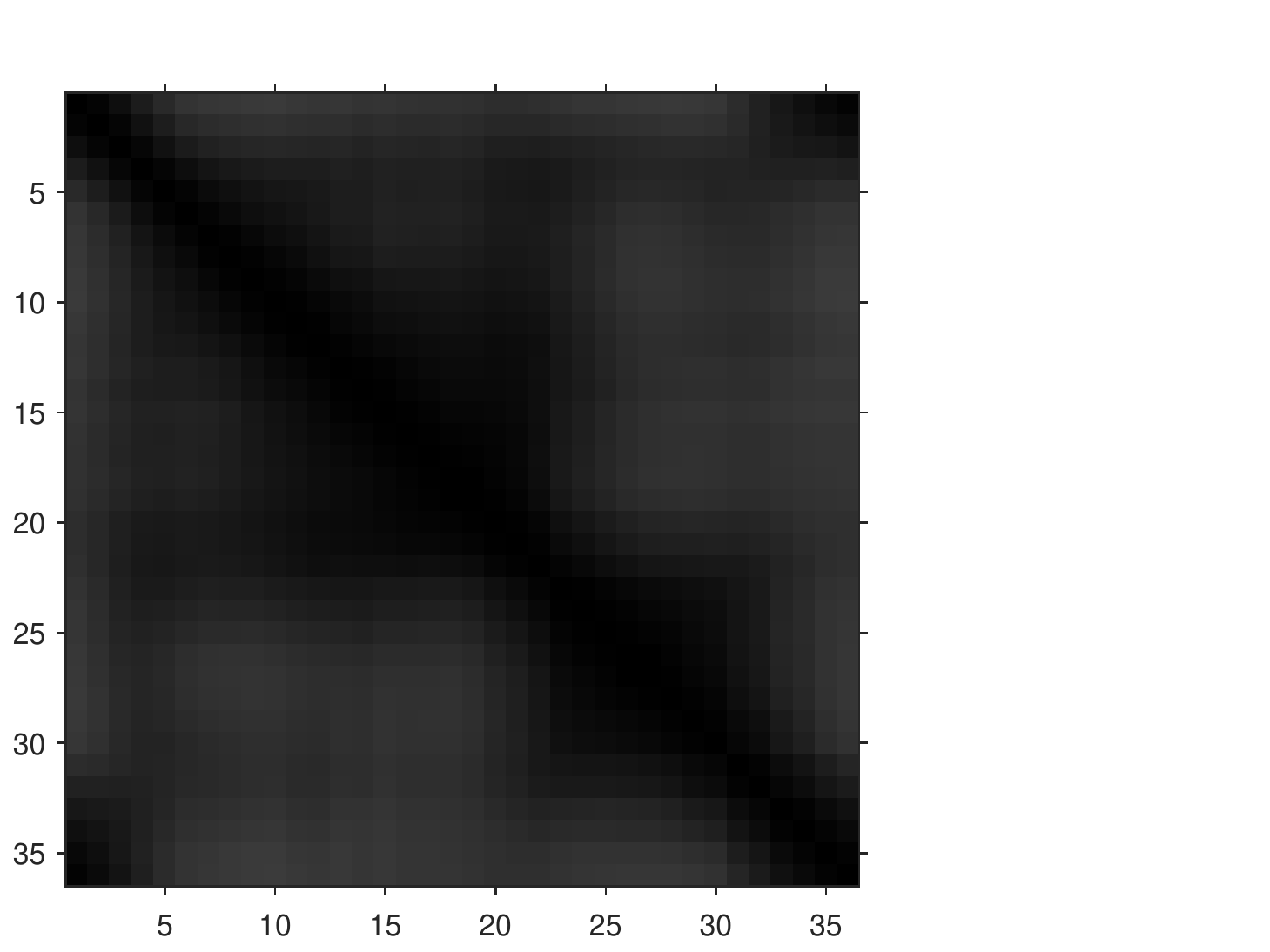}}
\subfigure[geometry distance]{\includegraphics[width=0.2\linewidth, bb=0 0 290 290]{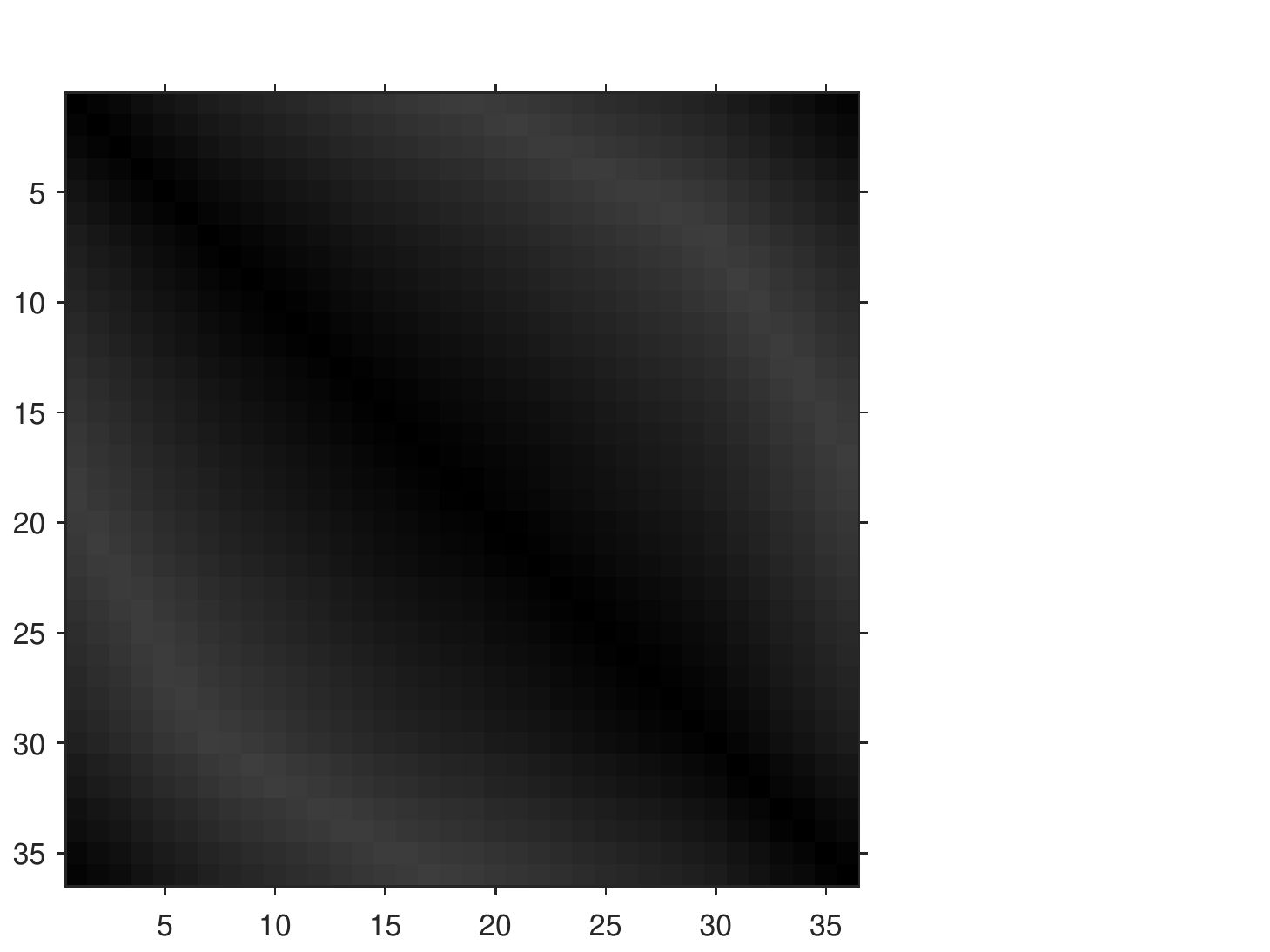}}
\subfigure[GCW-SSIM distance (50x)]{\includegraphics[width=0.2\linewidth, bb=0 0 290 290]{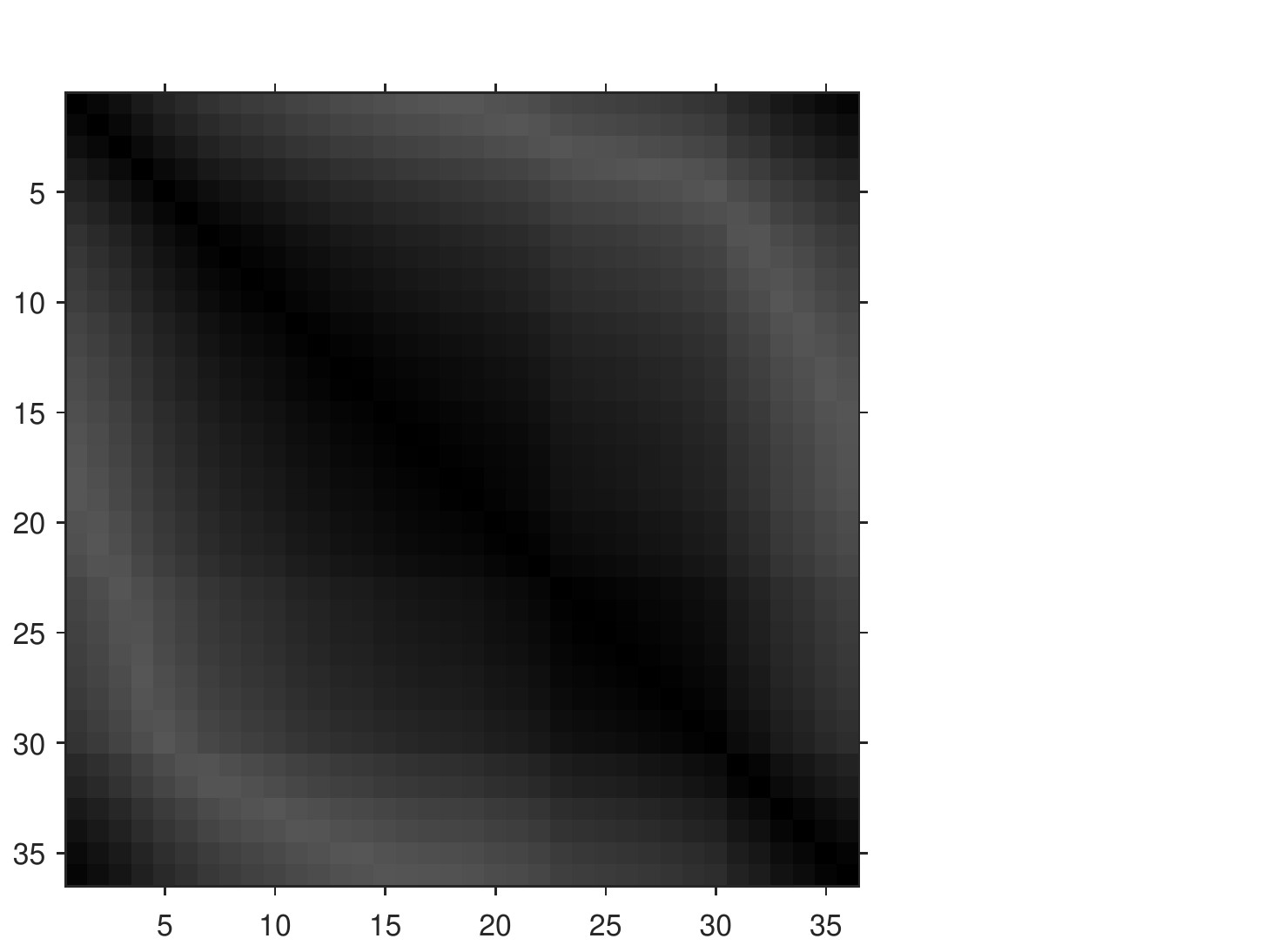}}
\caption{Visualization of the distances on the toy cat set (1 unit = 10 degrees rotation).
The distances are normalized into the range $[0,1]$ for visualization by dividing the maximum distance.
Note that the GCW-SSIM distance is enlarged by a factor of $50$ for showing due to their small values.}
\label{fig:toy-cat-dist}
\end{figure*}

The k-medoids algorithm is a variant of the $k$-means clustering algorithm. It selects data samples as
centers (also called \textit{medoids}) and attempts to minimize the following objective function:
\begin{equation}
\bm{C}^{*}=\mathop{\arg\min}\limits_{\bm{C}}\sum_{i=1}^{m}\sum_{j=1}^{k}v_{ij}\,d(\bm{x}^{(i)},\bm{c}^{(j)}) \label{eq:k-medoids}
\end{equation}
where $v_{ij}$ is the assignment index: $v_{ij}$ equals $1$ if $\bm{x}^{(i)}$ is assigned to the $j$th cluster
or $0$ otherwise. $d(\bm{x}^{(i)},\bm{c}^{(j)})$ is any kind of distance measures. The most famous algorithm
to solve (\ref{eq:k-medoids}) is the Partitioning Around Medoids (PAM) algorithm \cite{bib:pam} which starts
from some randomly selected medoids and iteratively updates the assignments and medoids until the objective
function achieves some local minima.

We use the GCW-SSIM distance proposed in the previous section as the distance measure for the $k$-medoids
algorithm. The corresponding clustering algorithm is named as the GCW-SSIM $k$-medoids. The procedure for
GCW-SSIM $k$-medoids is presented in Algorithm \ref{alg:gcw-ssim-k-medoids}.
Here $g(\bm{x}, \bm{c})$ denotes the geometric CW-SSIM distance between $\bm{x}$ and $\bm{c}$.
The GCW-SSIM $k$-medoids mainly deals with the the unsupervised clustering task of visual objects that lies on a manifold.

\begin{algorithm}[!htb]
\caption{The GCW-SSIM $k$-medoids Algorithm}
\label{alg:gcw-ssim-k-medoids}
\begin{algorithmic}[1]
\REQUIRE $\bm{X}$ and $t$
\ENSURE $\bm{C}^*$ (optimal medoids)
\STATE $\bm{C}\leftarrow \bm{C}^{(0)}$ (randomly selected $k$ medoids)
\STATE $\bm{G}\leftarrow $ GCW-SSIM distance computed by Algorithm \ref{alg:gcw-ssim}
\WHILE {there are changes in the assignments}
\STATE Assign each data point to the closest medoid;
\begin{eqnarray}
v_{ij} \leftarrow
\begin{cases}
1, & \text{if~} g_{ij} = \mathop{\min}\limits_{j=1,2,\cdots,k} g(\bm{x}^{(i)}, \bm{c}^{(j)}); \\
0, & \text{otherwise}.
\end{cases}
\end{eqnarray}
\STATE Update medoids as follows,
$$\bm{c}^{(j)}\leftarrow\mathop{\arg\min}\limits_{\bm{c}\in\{\bm{x}^{(i)}|v_{ij}=1\}}
\sum_{i:v_{ij}=1} g(\bm{x}^{(i)},\bm{c})$$
\ENDWHILE
\STATE $\bm{C}^* \leftarrow \bm{C}$
\RETURN $\bm{C}^*$
\end{algorithmic}
\end{algorithm}

\begin{table*}[!thb]
\caption{Clustering performance comparison on the Coil-sets.}
\label{tab:clustering-coil-sets}
\centering
\begin{tabular}{|c!{\vrule width 0.8pt}c|c|c!{\vrule width 0.8pt}c|c|c!{\vrule width 0.8pt}c|c|c!{\vrule width 0.8pt}c|c|c|}  \hline
\multirow{2}{*}{Data Sets} & \multicolumn{3}{c!{\vrule width 0.8pt}}{$k$-medoids ($L_2$)} & \multicolumn{3}{c!{\vrule width 0.8pt}}{$k$-medoids (C)} & \multicolumn{3}{c!{\vrule width 0.8pt}}{$k$-medoids (G)} & \multicolumn{3}{c|}{$k$-medoids (GC)} \\ \cline{2-13}
& $r_{e}$ & $r_{t}$ & $r_{f}$ & $r_{e}$ & $r_{t}$ & $r_{f}$  & $r_{e}$  & $r_{t}$ & $r_{f}$ & $r_{e}$ & $r_{t}$ & $r_{f}$ \\ \Xhline{0.8pt}
Coil-5  & 41.7 & 50.0 & 15.8 & 43.0 & 58.0 & 21.9 & ~9.2 & 89.9 & ~4.6 & \textbf{~2.5} & \textbf{95.6} & \textbf{~1.3} \\ \hline
Coil-10 & 24.9 & 80.5 & ~5.5 & 23.2 & 74.3 & ~5.4 & ~8.8 & 89.5 & ~1.4 & \textbf{~0.3} & \textbf{99.5} & \textbf{~0.1} \\ \hline
Coil-15 & 30.7 & 71.3 & ~4.1 & 32.4 & 63.7 & ~4.5 & 10.9 & 91.2 & ~1.6 & \textbf{~9.2} & \textbf{93.1} & \textbf{~1.3} \\ \hline
Coil-20 & 33.2 & 60.4 & ~2.7 & 36.4 & 62.4 & ~4.7 & 19.4 & 79.3 & \textbf{~1.8} & \textbf{15.8} & \textbf{87.2} & ~2.0\\ \hline
\end{tabular}
\end{table*}

\section{Experiments} \label{sec:experiments}

In this section, experiments are conducted to evaluate the performance of the proposed GCW-SSIM distance and
the GCW-SSIM $k$-medoids. The following three criteria are used to evaluate the performances of unsupervised image
categorization \cite{bib:affinity-propagation}.
1) Each learned category is associated with the true category that accounts for the largest number of cases in the
learned category. Thus, the \textit{error rate} ($r_{e}$) is computed.
2) \textit{Rate of true association} ($r_{t}$) is the fraction of pairs of images from the same true category
that was correctly placed in the same learned category.
3) \textit{Rate of false association} ($r_{f}$) is the fraction of pairs of images from different true categories
that was erroneously placed in the same learned category. Better clustering performance is characterized with
lower values for $r_{e}$ and $r_{f}$ but higher $r_t$ value.

The $k$-medoids is repeated $1000$ times to reduce the affect of randomly selected initial seeds.
The criterion values are recorded when the objective function (\ref{eq:k-medoids}) achieves a minimum value
among the repeats.

All experiments are conducted on a single PC with Intel i7-4770 CPU (4 Cores) and 16G RAM.

\begin{table*}[!htb]
\caption{Clustering performance comparison on the large Coil-sets. Definition of $r_{e}$, $r_{t}$ and $r_{f}$
can be found in Sec. \ref{sec:experiments}.}\label{tab:clustering-large-coil-sets}
\centering
\begin{tabular}{|c!{\vrule width 0.8pt}c|c|c!{\vrule width 0.8pt}c|c|c!{\vrule width 0.8pt}c|c|c!{\vrule width 0.8pt}c|c|c|}  \hline
\multirow{2}{*}{Data Sets} & \multicolumn{3}{c!{\vrule width 0.8pt}}{$k$-medoids ($L_2$)} & \multicolumn{3}{c!{\vrule width 0.8pt}}{$k$-medoids (C)} & \multicolumn{3}{c!{\vrule width 0.8pt}}{$k$-medoids (G)} & \multicolumn{3}{c|}{$k$-medoids (GC)} \\ \cline{2-13}
& $r_{e}$ & $r_{t}$ & $r_{f}$ & $r_{e}$ & $r_{t}$ & $r_{f}$  & $r_{e}$  & $r_{t}$ & $r_{f}$ & $r_{e}$ & $r_{t}$ & $r_{f}$ \\ \Xhline{0.8pt}
Coil-25  & 41.9 & 59.8 & ~4.2 & 38.2 & 55.7 & ~2.7 & 25.3 & 72.6 & ~2.4 & \textbf{20.1} & \textbf{77.7} & \textbf{~1.5} \\ \hline
Coil-50  & 42.6 & 54.6 & ~2.3 & 41.0 & 56.5 & ~1.9 & 32.6 & 70.0 & ~1.6 & \textbf{24.6} & \textbf{78.1} & \textbf{~1.2} \\ \hline
Coil-75  & 48.0 & 47.8 & ~1.4 & 50.5 & 49.4 & ~1.5 & 36.3 & 68.2 & ~1.8 & \textbf{32.4} & \textbf{69.8} & \textbf{~1.0} \\ \hline
Coil-100 & 52.6 & 47.1 & ~1.4 & 51.9 & 44.3 & ~1.1 & 37.3 & 62.9 & ~1.1 & \textbf{35.4} & \textbf{67.2} & \textbf{~1.0} \\ \hline
\end{tabular}
\end{table*}

\begin{figure*}[!htb]
\centering
\subfigure[$k$\,--\,$r_{e}$ graph]{\includegraphics[width=0.3\linewidth]{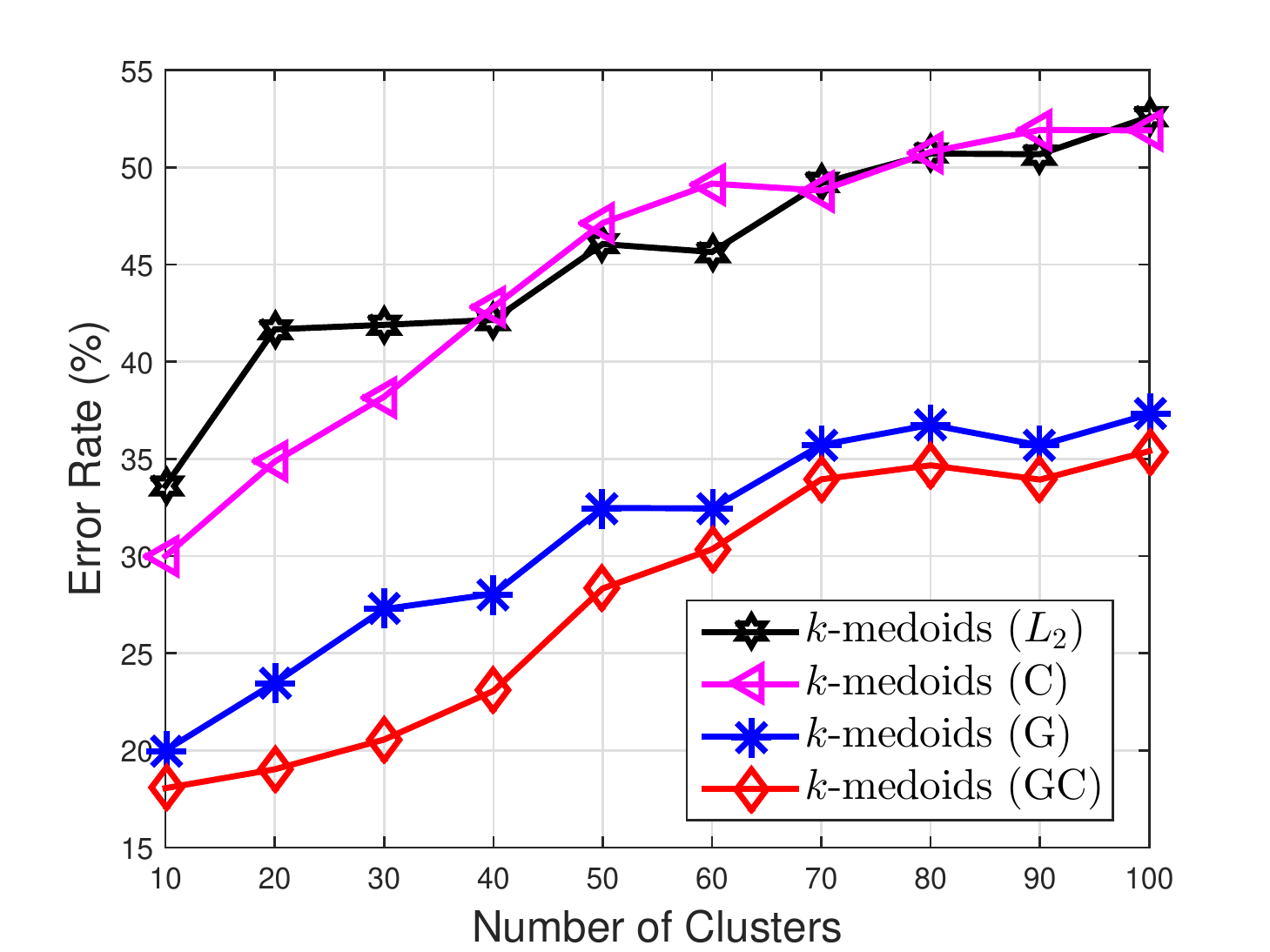}}
\subfigure[$k$\,--\,$r_{t}$ graph]{\includegraphics[width=0.3\linewidth]{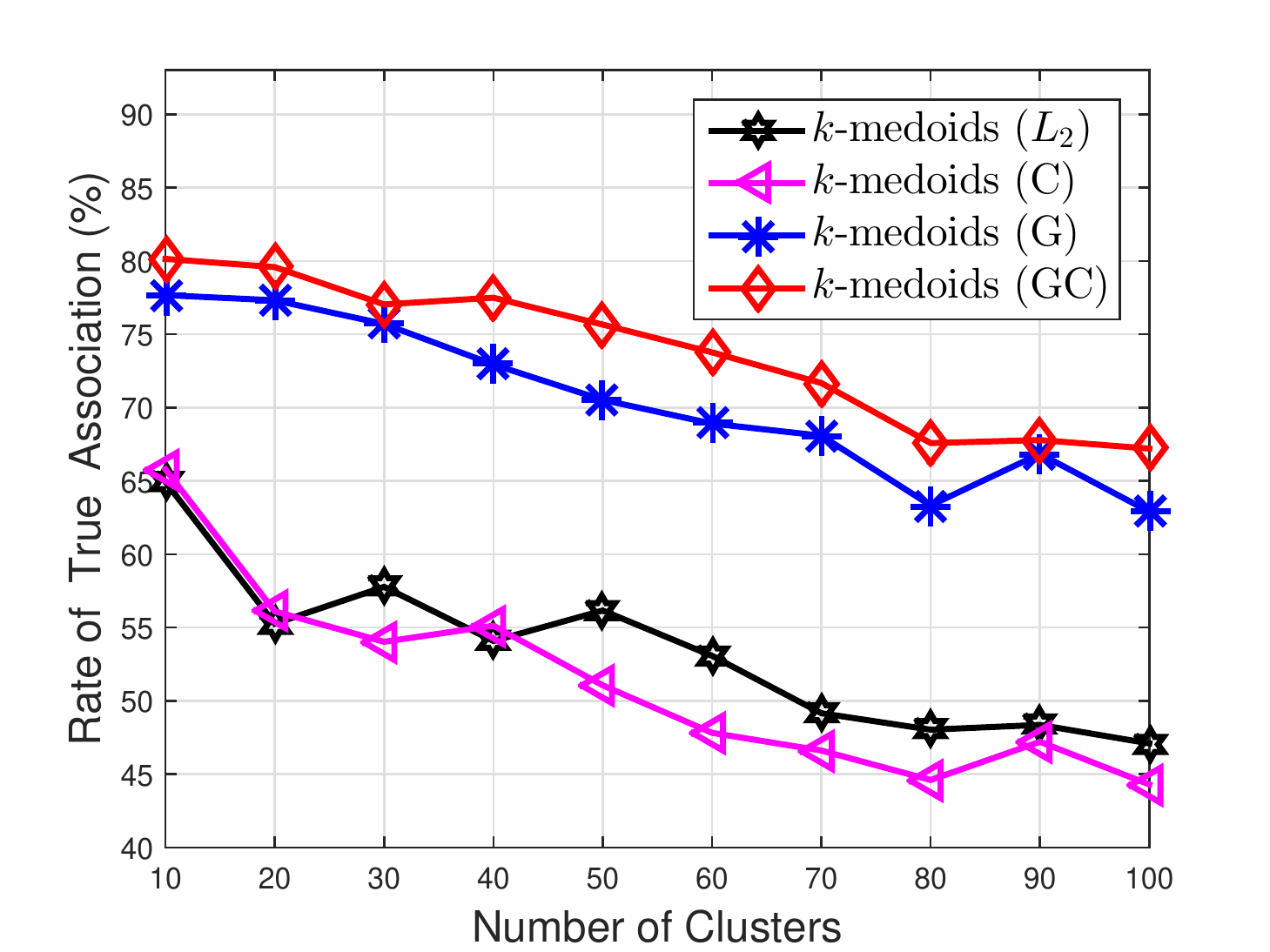}}
\subfigure[$k$\,--\,$r_{f}$ graph]{\includegraphics[width=0.3\linewidth]{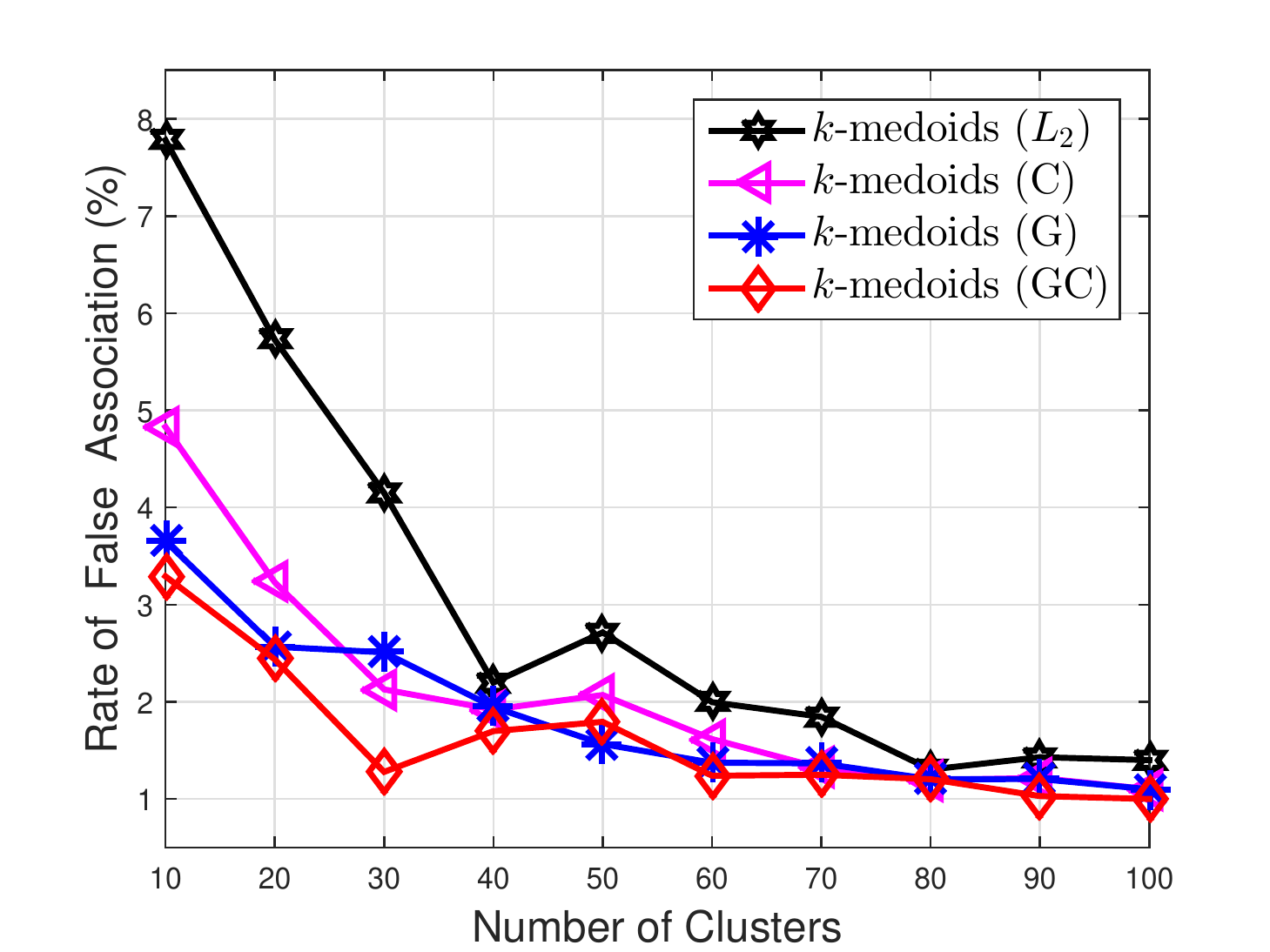}}
\caption{Performance comparison on the large Coil-100 dataset with a varying number of clusters.}
  \label{fig:coil100-performance}
\end{figure*}

\subsection{Experiments on Coil-20}

We first shows the difference of distances on the toy cat set (the 4th category in Coil-20). The toy cat set
contains $72$ images taking $5$ degrees apart as the object rotated on a turntable. Fig. \ref{fig:toy-cat-dist}
shows the differences of the four distances: 1) The $L_2$ distance gets small value only when the difference
of angles between two objects is no more than $20$ degrees; 2) the CW-SSIM distance could get a small value
even when the difference of angles is more than $100$ degrees; 3) The geometry distance narrows the distance
gaps in $L_2$ since the objects are locally connected; and 4) the GCW-SSIM distances are relatively smaller
than the geometry distance, and as a result it is inclined to cluster the objects into the same category.
This means that the GCW-SSIM distance would be more distinctive in clustering the toy cat into the same
category than the other three distances.

We now compare the clustering performance of the $k$-medoids on the dataset Coil-20 with the four distance
measures. We use ``C'' to denote the CW-SSIM distance, ``G'' to represent the geometric distance and ``GC''
to denote the GCW-SSIM distance. Three subsets are selected from the dataset Coil-20.
They are Coil-5 (object 1,3,5,7 and 9), Coil-10 (objects with even numbers) and Coil-15
(objects except 3,7,11,15 and 19). The clustering results on the Coil-sets are shown in
TABLE \ref{tab:clustering-coil-sets}. As expected, the GCW-SSIM $k$-medoids gets the best results
compared with the other methods.

\subsection{Experiments on Coil-100}

In addition, we compare the clustering performance on Coil-100 to evaluate the effectiveness of the
proposed method with a large number of clusters. We also select three subsets from Coil-100.
They are Coil-25 (object 1, 5, 9, $\cdots$, 93 and 97), Coil-50 (objects with even number) and
Coil-75 (Coil-25 + Coil-50). Coil-25, Coil-50, Coil-75 and Coil-100 are denoted as the large Coil-sets
to distinguish from the Coil-sets. The clustering results on the large Coil-sets are shown in
TABLE \ref{tab:clustering-large-coil-sets}. Again, our method outperforms the other three methods.

\begin{table*}[htb]
\caption{Clustering performance comparison on the Olivetti-sets.}\label{tab:clustering-olivetti-sets}
\centering
\begin{tabular}{|c!{\vrule width 0.8pt}c|c|c!{\vrule width 0.8pt}c|c|c!{\vrule width 0.8pt}c|c|c!{\vrule width 0.8pt}c|c|c|}  \hline
\multirow{2}{*}{Data Sets} & \multicolumn{3}{c!{\vrule width 0.8pt}}{$k$-medoids ($L_2$)} & \multicolumn{3}{c!{\vrule width 0.8pt}}{$k$-medoids (C)} & \multicolumn{3}{c!{\vrule width 0.8pt}}{$k$-medoids (G)} & \multicolumn{3}{c|}{$k$-medoids (GC)} \\ \cline{2-13}
& $r_{e}$ & $r_{t}$ & $r_{f}$ & $r_{e}$ & $r_{t}$ & $r_{f}$  & $r_{e}$  & $r_{t}$ & $r_{f}$ & $r_{e}$ & $r_{t}$ & $r_{f}$ \\ \Xhline{0.8pt}
Oliv.-10 & ~5.0 & 90.9 & ~1.3 & 15.0 & 77.1 & ~2.9 & \textbf{~4.1} & 92.4 & \textbf{~1.0} & 11.0 & \textbf{93.6} & ~2.4 \\ \hline
Oliv.-20 & 36.5 & 52.6 & ~4.0 & 33.5 & 54.1 & \textbf{~3.1} & 34.5 & 57.0 & ~3.3 & \textbf{30.0} & \textbf{73.2} & ~3.7 \\ \hline
Oliv.-30 & 34.7 & 53.0 & ~2.4 & 34.7 & 54.7 & ~2.3 & 28.0 & 65.4 & ~2.0 & \textbf{25.0} & \textbf{71.8} & \textbf{~1.7} \\ \hline
Oliv.-40 & 39.5 & 47.0 & ~2.1 & 36.3 & 50.2 & \textbf{~1.5} & 34.8 & 57.6 & ~1.9 & \textbf{29.7} & \textbf{69.2} & ~2.6 \\ \hline
\end{tabular}
\end{table*}

To get a closer look at the tendency of the criteria with a variant number of clusters, we select the first $k$
($k$ = 10, 20, $\cdots$, 90 and 100) categories from the Coil-100 set as $10$ subsets. The performance results are
visualized in Fig. \ref{fig:coil100-performance}. It is observed from Fig. \ref{fig:coil100-performance}(a)-(b)
that the manifold distance based methods outperform the non-manifold distance based methods.
It is observed further that GCW-SSIM $k$-medoids outperforms $k$-medoids with the geometric distance.

\subsection{Experiments on Olivetti Face Database}

We now compare the clustering performance with different distances on the Olivetti Face Database for face recognition.
The data set consists of $400$ face images from $40$ individuals. The images are taken at different times,
with varying lighting, facial expressions and facial details. The size of each image is $64\times 64$ pixels.
We select three subsets, Oliv.-10 (faces 2,6,10, $\cdots$, 34 and 38), Oliv.-20 (faces with odd number),
and Oliv.-30 (Oliv.-10 + Oliv.-20). The whole database is denoted as Oliv.-40.

The results are shown in TABLE \ref{tab:clustering-olivetti-sets}. Our method gets the best performance on
most of the subsets under the criteria $r_{e}$ and $r_{t}$. This implies that the GCW-SSIM distance is less
sensitive to lighting and facial expression changes. As a result, the GCW-SSIM distance makes faces in the same
true category more similar compared with the other distances and thus helps to cluster the faces into a same
category. In addition, the $r_{t}$ value of GCW-SSIM $k$-medoids on the Oliv.-40 data set outperforms
the state-of-art performance reported in \cite{bib:alex2014}, which is around $68\%.$

\section{Conclusion}\label{sec:conclusion}

In this paper, we proposed a new distance measure named GCW-SSIM distance which has the merit of both
the CW-SSIM index and the geometric distance. Compared with the geometric distance, the GCW-SSIM distance
improves the accuracy of computing the manifold distance since the CW-SSIM index is more robust to rotations
than the $L_2$ distance. In addition, the GCW-SSIM distance is more robust to rotations and translations of
images compared with the CW-SSIM index, which is verified by visualization on the toy cat set. We also proposed
a new clustering method named GCW-SSIM $k$-medoids that uses the GCW-SSIM distance for visual object clustering.
The experiments on the real world data sets showed that GCW-SSIM $k$-medoids has an excellent performance
for the visual object categorization tasks.


\IEEEtriggeratref{13}

\end{document}